\documentclass[runningheads]{llncs}
\usepackage[T1]{fontenc}
\usepackage{graphicx}
\usepackage{CJK}

\usepackage{float}
\usepackage{url}
\usepackage{enumitem}
\usepackage{todonotes}
\usepackage{booktabs}
\usepackage{dashrule}
\usepackage{mdframed}
\usepackage{listings}
\usepackage[normalem]{ulem}
\usepackage{subcaption}
\usepackage{tikz}
\usetikzlibrary{positioning}
\usepackage{xcolor}
\usepackage{setspace}
\usepackage{amsmath}

\begin{document}
\begin{CJK*}{UTF8}{gbsn}
\title{ThinknCheck: Grounded Claim Verification with \\ Compact, Reasoning-Driven, and Interpretable Models}
\titlerunning{ThinknCheck}

\author{Delip Rao\orcidID{0000-0002-4534-9906} \and
Feijiang Han\orcidID{0009-0009-3636-1012} \and
Chris Callison-Burch\orcidID{0000-0001-8196-1943}}

\authorrunning{D. Rao et al.}

%
\institute{University of Pennsylvania, Philadelphia, PA, USA \\
\email{\{delip, feijhan, ccb\}@seas.upenn.edu}}

\maketitle              

\begin{abstract}
We present ThinknCheck, a 1B-parameter verifier for grounded claim verification that first produces a short, structured rationale and then a binary verdict. We construct LLMAggreFact-Think, a 24.1k reasoning-augmented training set derived from LLMAggreFact, and fine-tune a 4-bit Gemma3 model to follow this format. On LLMAggreFact, ThinknCheck attains 78.1 balanced accuracy (BAcc), surpassing MiniCheck-7B (77.4) with 7x fewer parameters; removing the reasoning step reduces BAcc to 57.5. On SciFact, ThinknCheck reaches 64.7 BAcc, a +14.7 absolute gain over MiniCheck-7B. By contrast, zero-shot chain-of-thought on the base Gemma3-1B harms accuracy relative to direct answers, and preference optimization with a simple format+accuracy reward underperforms supervised reasoning. To probe the latter, we introduce GSMClaims and a domain-specialized variant, ThinknCheck-Science, which improves across benchmarks, including 61.0\% accuracy on GSMClaims. Overall, explicit, supervised reasoning enables compact verifiers that are competitive while remaining resource-efficient and interpretable.

\end{abstract}
\section{Introduction}
\label{sec:introduction}
Large Language Models (LLMs) are increasingly pivotal in modern AI-driven workflows involving natural language claim verification \cite{thorne2018fever}, yet their propensity for hallucinations \cite{zhang2023language}, reasoning errors \cite{chen2023understanding}, and their opaque nature limit broader deployment, particularly in high-stakes domains like healthcare and scientific innovation.

\begin{figure*}[t!]
    \centering
    \includegraphics[width=1\linewidth]{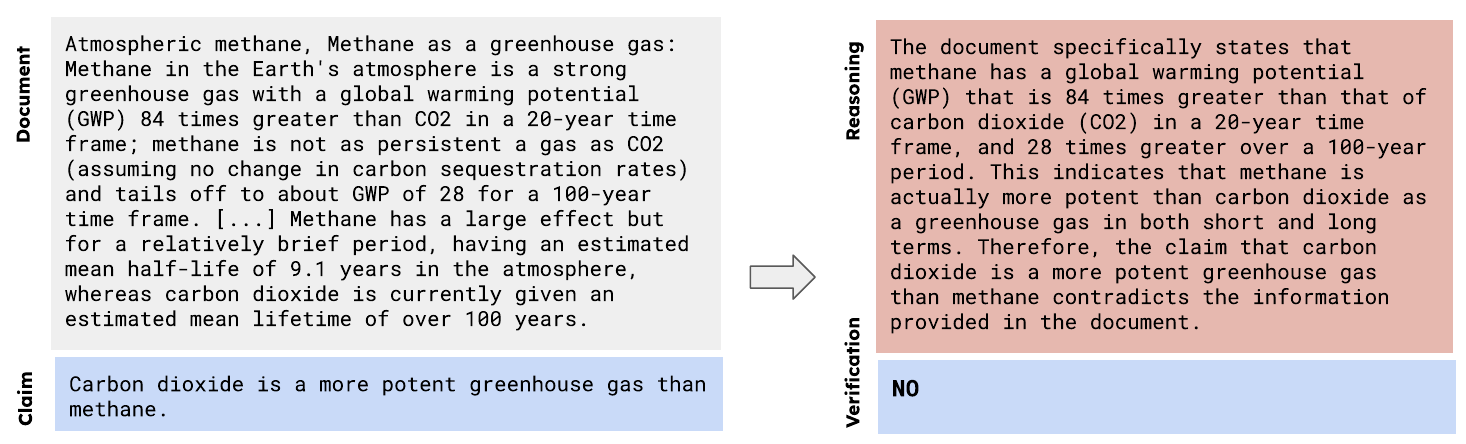}
    \vspace{-1.2em}
        \caption{A sample from the LLMAggreFact-Think dataset, which also illustrates our formulation of the claim verification task: Given a pair of claim and document, our goal is to produce cogent reasoning in addition to the verification label. The \texttt{[...]} represents parts of the reasoning tokens that we elided to accommodate the example in this figure.}
        \vspace{-1.0em}
    \label{fig:llmaggrefact-think-example}
\end{figure*}

Current claim verification approaches, while evolving, often face challenges. Some methods can be computationally intensive, requiring multiple LLM calls for a single verification \cite{malaviya2024expertqa,jacovi2024chainofthought}. Verification using large, closed-source models also raises concerns regarding cost, privacy, and data security. While the trend in general reasoning models has seen the development of very large systems, such as OpenAI's o-series and DeepSeek's R1 \cite{deepseek-r1-2025}, which aim for broad reasoning capabilities, there is a concurrent need for smaller, more specialized models \cite{tang2024minicheck} that can perform robustly on specific tasks like claim verification, especially in resource-constrained environments. Our work aligns with this latter direction, focusing on creating efficient yet powerful verification models.

To address these challenges, we introduce ThinknCheck, a suite of novel low-footprint claim verification models that explicitly generate structured reasoning chains \textit{before} rendering a verification decision. Specifically ThinknCheck is a 4-bit quantized 1B parameter Gemma3 \cite{gemmateam2025gemma3technicalreport} model, fine-tuned on our newly createdLLMAggreFact-Think dataset---a version of theLLMAggreFact benchmark \cite{tang2024minicheck} that we augmented with explicit reasoning traces. As illustrated in Figure~\ref{fig:llmaggrefact-think-example}, explicitly generating reasoning allows ThinknCheck to handle claims that require multi-step inference, a common scenario where prior models falter. Our contributions are as follows:
\vspace{-0.1em}
\begin{itemize}[noitemsep,topsep=0pt]
    \item Introduce \textsc{ThinknCheck}, a \textbf{reasoning-optimized document-grounded claim verifier} that reasons before verifying, improving accuracy and interpretability by producing a concise rationale before the decision.
    \item Demonstrate ThinknCheck's explicit reasoning significantly \textbf{boosts verification accuracy} (+20.6 points over non-reasoning ablation) and \textbf{substantially improves out-of-domain generalization} (+14.7 points on scientific claims).
    \item Create and release GSMClaims, a novel benchmark from reformulated grade school math problems, to \textbf{evaluate arithmetic reasoning capabilities in claim verification systems}.
    \item Develop ThinknCheck-Science, a specialized variant \textbf{optimized for scientific and mathematical verification}, achieving significant performance improvements across relevant benchmarks.
    \item Answer how much ``thinking'' is optimal for small models: To demystify this, we empirically relate rationale length to verification accuracy. We observe an inverted-U pattern: mid-length reasoning performs best, while \textbf{very short and very long chains degrade performance in distinct ways} (See Fig.~\ref{fig:length-bacc}).
    \item \textbf{Open source all} created datasets and models under an Apache 2.0 license\footnote{URL withheld for blind-review}.
\end{itemize}

\section{Related Work}
\label{sec:related_work}

Our work connects four threads: (i) claim verification and benchmarks, (ii) reasoning in LMs, (iii) lightweight verifiers, and (iv) supervision for reasoning.
\vspace{-0.5em}
\paragraph{Claim verification and benchmarks.}
Early resources such as \textsc{FEVER} and \textsc{SciFact} established document-grounded claim verification as a supervised classification task mapping a (claim, document) pair to a support decision \cite{thorne2018fever,wadden-etal-2020-fact,tang2024minicheck}. Hallucination and opacity concerns with LLMs motivate verifiers that are both accurate and interpretable \cite{chen2023understanding,zhang2023language}. \textsc{LLMAggreFact} aggregates nine sources to test diverse scenarios \cite{tang2022understanding,nallapati2016abstractive,narayan2018don,zhu2021mediasum,hu2023meetingbank,liu2023evaluating,malaviya2023expertqa,wang2023factcheck,kamoi2023wice}, while domain-specific suites such as \textsc{SciFact} probe scientific claims and \textsc{GSM8K} targets math reasoning \cite{wadden-etal-2020-fact,cobbe2021}. To directly assess numerical reasoning in verification, we reformulate \textsc{GSM8K} into \textsc{GSMClaims}.
\vspace{-0.5em}
\paragraph{Reasoning-augmented verification.}
Reasoning traces (e.g., Chain-of-Thought) can help models explain intermediate steps \cite{wei2022chain} and have been adapted via verifiable-CoT and ReAct to interleave reasoning with actions \cite{jacovi2024chain,yao2023react}. While very large models (e.g., DeepSeek R1) aim for broad reasoning \cite{deepseek-r1-2025}, multiple studies find that vanilla CoT can underperform at small scales \cite{li2025smallmodelsstrugglelearn,liu2025mindstepbystep}. We fine-tune a compact model to produce structured, pre-decision rationales specific to claim verification, and observe that preference-optimization methods such as GRPO---which have shown promise at larger scales \cite{shao2024deepseekmathpushinglimitsmathematical,liu2025understandingr1zeroliketrainingcritical,zheng2025groupsequencepolicyoptimization}---do not help at 1B parameters.
\vspace{-0.5em}
\paragraph{Lightweight and specialized verifiers.}
There is growing interest in small, efficient models for deployment \cite{allal2025smollm2}. \texttt{MiniCheck} (7B) demonstrated the value of a purpose-built verifier trained on synthetic data and outperformed \texttt{AlignScore} on LLMAggreFact \cite{tang2024minicheck,zha2023alignscore}. However, \texttt{MiniCheck} struggles on multi-step cases and does not produce explanations that are important for trust \cite{Bansal2020DoesTW,Fan2021HumanAICF,Javaid2021ExplanationsFA}. ThinknCheck is a 1B model that \emph{reasons before deciding}, yielding concise rationales with lower footprint.
\vspace{-0.5em}
\paragraph{Supervising reasoning.}
We supervise rationales via SFT on synthetic traces (knowledge distillation) \cite{xu2024survey}. Although preference optimization (e.g., GRPO) is another path \cite{shao2024deepseekmath,zheng2025groupsequencepolicyoptimization,liu2025understandingr1zeroliketrainingcritical}, in our 1B setting SFT on structured reasoning yields better outcomes. We use a standard TRL stack for reproducibility \cite{vonwerra2022trl}.

\section{Problem Formulation}
\label{sec:background}
The standard formulation of document-grounded claim verification, as used by previous research~\cite{tang2024minicheck} and predecessors, is a classification task: a discriminator $\mathcal{M}$ maps a claim (from space $\mathcal{C}$) and document (from space $\mathcal{D}$) to a discrete label in $\{0, 1\}$ ($1$ for supported, $0$ otherwise): $\mathcal{M}: \mathcal{C} \times \mathcal{D} \rightarrow \{0, 1\}$

Our work extends this by incorporating explicit reasoning. We define this task with a reasoner $\mathcal{R}$ that maps the input claim-document pair to a reasoning trace $\mathcal{T}$ and a boolean verification label: $\mathcal{R}: \mathcal{C} \times \mathcal{D} \rightarrow \mathcal{T} \times \{0, 1\}$
This richer output format enhances interpretability and aims to improve accuracy by requiring the model to articulate its reasoning. We adopt the binary labels \textsc{Supported (1)} and \textsc{NotSupported (0)} from prior work, treating ``\textsc{Refutes}'' and ``\textsc{NotSupported}'' identically\footnote{We concur with previous research~\cite{tang2024minicheck} that ``\textsc{Refutes}'', common in general NLI problems, is rare in claim verification.}.

\section{Dataset and Model Development}
\label{sec:methodology}
This section details the creation of our training dataset LLMAggreFact-Think and the model training procedures for ThinknCheck.

\subsection{LLMAggreFact-Think Dataset Construction}
\label{sec:llmaggrefact-think}
To train our reasoning-based verifier, we created LLMAggreFact-Think by augmenting the 30.4K examples in the LLMAggreFact development set with reasoning chains. Using zero-shot prompting, GPT-4o-mini\footnote{Accessed on March 4, 2025. We did not use the o-series models for this as it does not provide access to raw reasoning tokens.} generated a step-by-step reasoning process and a YES/NO verification label for each (document, claim) pair --- see Figure \ref{fig:llmaggrefact-think-example}; prompt in Appendix \ref{appendix:think-prompt}. For high-quality reasoning, we filtered instances where GPT-4o-mini's generated label mismatched the original LLMAggreFact label, reducing the dataset from 30.4K to 24.1K examples\footnote{Notably, $\sim$21\% of LLMAggreFact dev set labels differed from GPT-4o-mini's predictions; analyzing this discrepancy is beyond this paper's scope. Hence we chose to only train on examples with agreement.}. This filtered set, LLMAggreFact-Think, contains 4-tuples: (claim, document, verification label, reasoning). We opted against using reasoning traces from \texttt{Deepseek R1} \cite{deepseek-r1-2025} due to their verbosity and token inefficiency. To ensure quality, we randomly sampled 100 samples across all 9 datasets in LLMAggreFact and manually inspected the reasoning traces derived from GPT-4o and found them to be accurate.

\subsection{ThinknCheck-1B Model Training}
\label{sec:thinkncheck-training}
We implemented ThinknCheck-1B by fine-tuning a 4-bit quantized Gemma3 1B model on LLMAggreFact-Think (training details in Appendix \ref{appendix:finetune-hyperparms}). Our choice of Gemma3 was inspired by its size, recency, and its overall performance across diverse LLM benchmarks \cite{gemmateam2025gemma3technicalreport}.
The fine-tuning prompt (Appendix \ref{appendix:finetune-prompt-thinkncheck}) mirrored the LLMAggreFact-Think data structure, constraining the model to output both reasoning and the final verification solution.\footnote{Inference uses  parameters recommended by the Gemma3 paper: temperature=1.0, top\_p=0.95, and top\_k=64.} For finetuning we used the 24.1K examples constructed as reported in Section~\ref{sec:llmaggrefact-think} and retained 20\% of that as a development set.

\subsection{Ablation Model Variants: Base model, Non-thinking, CoT, and GRPO}
\label{sec:model-variants}
To isolate the reasoning step's impact, we trained an ablation model, ThinknCheck-nothink-1B. It shares ThinknCheck-1B's architecture, data, and hyperparameters but was trained with a prompt (Appendix \ref{appendix:finetuning-prompt-nothink}) requesting only the final solution, omitting reasoning generation. This ablation ensures that observed performance gains are not solely due to our choice of Gemma3 as the backbone. For context, we also evaluate the Gemma3-1B base model without fine-tuning (direct YES/NO) and with a Chain-of-Thought (CoT) prompt~\cite{wei2022chain} at inference time using the same decoding settings; these rows appear in Table~\ref{tab:main-results}. Beyond SFT, we train a preference-optimized variant using the Dr. GRPO variant with GSPO enabled in HuggingFace's TRL library~\cite{vonwerra2022trl}. We experimented with LoRA adapters initialized with and without SFT warm start, each jointly optimizing a two-term reward: (i) a \emph{format-adherence} bonus that requires the model to emit both \texttt{<REASONING>} and \texttt{<SOLUTION>} in order; and (ii) a \emph{class-weighted accuracy} term that upweights the minority class to counter label imbalance. \footnote{We keep 4‑bit loading and the same LoRA configuration as in Section~\ref{sec:thinkncheck-training}, generate $5$ candidates per prompt, and train for a single epoch (400 steps) with fused AdamW ($5{\times}10^{-6}$ LR), warmup ratio $0.1$, per-device batch size $4$ and gradient accumulation $4$.} Full reward definitions, numeric weights, sampling limits, logging/checkpointing, and other reproducibility details are provided in Appendix~\ref{app:grpo}; empirical results are reported in Section~\ref{sec:core-ablations}.

\subsection{GSMClaims Dataset for Arithmetic Reasoning}
\label{sec:gsmclaims-creation}
We construct \textbf{GSMClaims} from the GSM8K test set \cite{cobbe2021} by reframing each problem into \emph{two} claim-verification instances using GPT-4o: (1) convert the problem context into a reference document; (2) generate a \emph{positive} claim whose answer is the correct computed value; and (3) generate a \emph{negative} claim with a plausible but incorrect value reflecting common calculation errors. This 3-step process yields \textbf{2{,}634} balanced items in which verification hinges on performing the arithmetic grounded in the document. We manually inspected a subset of 100 positive and negative claims and found them to be near-perfectly accurate; the prompt template is provided in Appendix~\ref{appendix:GSMClaims-prompt}.

\subsection{ThinknCheck-Science: Specializing for Complex Claims}
To better handle quantitative and scientific claims, we train \textbf{ThinknCheck-Science}, a targeted variant of ThinknCheck-1B. We augment the LLMAggreFact-Think training data with \textbf{614} reasoning-enhanced examples from SciFact \cite{wadden-etal-2020-fact} and \textbf{398} from GSMClaims, emphasizing specialized knowledge and arithmetic calculation. ThinknCheck-Science retains the base architecture, quantization, and training procedure of ThinknCheck-1B, differing only in the enriched supervision set aimed at strengthening numerical and domain-specific reasoning.

\section{Experiments}
\label{sec:experiments}

\noindent\textbf{Research Questions.} We structure the evaluation around three questions: \textbf{RQ1}. Do explicit, supervised rationales help at 1B scale? \textbf{RQ2}. Do they improve out-of-domain generalization (\textsc{SciFact}) over specialized baselines? \textbf{RQ3}. Do coarse preference rewards (GRPO) help or hurt?
\vspace{-0.5em}
\subsection{Evaluation Metrics}
Following prior work \cite{tang2022understanding,fabbri2021qafacteval,laban2022summac,tang2024minicheck}, we adopt Balanced Accuracy (BAcc) as our primary metric for evaluating potentially imbalanced datasets like LLMAggreFact and SciFact. \footnote{
\vspace{-0.5em}
{\tiny
$$\mathbf{BAcc = \frac{1}{2} \left( \frac{\text{TP}}{\text{TP}+\text{FN}} + \frac{\text{TN}}{\text{TN}+\text{FP}} \right),}$$
}
\vspace{-0.2em}
where TP, TN, FP, and FN represent true positives, true negatives, false positives, and false negatives, respectively.} For the balanced GSMClaims dataset, we report standard accuracy, so our results are interpretable with previous works.
\vspace{-0.5em}
\subsection{Baselines}
We compare ThinknCheck against three categories of baselines: (1) closed LLMs in zero-shot settings (GPT-4, GPT-4o, Claude-Sonnet-3.5) as reported by~\cite{tang2024minicheck}. All significance statements use 10k bootstrap resamples over the 29.3k LLMAggreFact test instances with paired resampling across models. Our goal is not to compete with these private and massive foundation models, but to provide context., (2) specialized verification models (AlignScore, MiniCheck-7B), and (3) ThinknCheck variants (ThinknCheck-nothink, ThinknCheck, ThinknCheck-Science) to isolate the impact of reasoning and data augmentation components.
\vspace{-0.5em}
\subsection{Core Verification Performance and Generalization}
\label{sec:core-ablations}

\begin{table}[h]
\centering
\small
\begin{tabular}{lcc}
\toprule
\textbf{Model} & \textbf{LLMAggreFact} & \textbf{SciFact} \\
               & \textbf{BAcc}         & \textbf{BAcc}    \\
\midrule
GPT-4 (zero-shot) & 75.3 & -- \\
GPT-4o (zero-shot) & 75.9 & -- \\
Claude-Sonnet-3.5 (zero-shot) & 77.2 & -- \\
\midrule
AlignScore (355M/fp16) & 70.4 & -- \\
MiniCheck (7B/fp16) & 77.4 & 50.0 \\
\midrule
Gemma3 (1B/fp4) (``base'') & 55.7 & -- \\
Gemma3 + CoT (1B/fp4) & 51.4 & -- \\
ThinknCheck-nothink (1B/fp4) & 57.5 & 21.7 \\
ThinknCheck (1B/fp4) & \textbf{78.1} & \textbf{64.7} \\
\bottomrule
\end{tabular}
\caption{Balanced accuracy (BAcc) on the \textsc{LLMAggreFact} test set (29.3k examples) and \textsc{SciFact} development set. The 1B \emph{reasoning-supervised} ThinknCheck model attains 78.1 BAcc on LLMAggreFact, exceeding the specialized 7B MiniCheck (77.4), and reaches 64.7 on SciFact (+14.7 over MiniCheck). Chain-of-Thought on the non-fine-tuned base reduces performance (55.7~$\rightarrow$~51.4), while answer-only SFT yields only a small gain (57.5). The ThinknCheck-nothink ablation scores 21.7 on SciFact, confirming that reasoning drives generalization. Significance: paired bootstrap, p~$<$~0.05.}
\label{tab:main-results}
\vspace{-0.75em}
\end{table}

Table~\ref{tab:main-results} summarizes results on LLMAggreFact and SciFact. The reasoning-supervised \textbf{ThinknCheck-1B} (4-bit) attains \textbf{78.1} BAcc on LLMAggreFact, surpassing MiniCheck-7B (77.4) despite using $\sim$7$\times$ fewer parameters, and matching or exceeding zero-shot closed models (GPT-4: 75.3; GPT-4o: 75.9; Claude Sonnet~3.5: 77.2). Reasoning is essential: removing it (\textit{ThinknCheck-nothink}) drops BAcc to 57.5, a $-20.6$ point decline. On the 1B base, zero-shot CoT hurts relative to direct answers (55.7 $\rightarrow$ 51.4), whereas supervised reasoning reverses the pattern and yields the best 1B result. All deltas are significant under paired bootstrap ($p<0.05$).

On SciFact, ThinknCheck-1B achieves 64.7 BAcc, a substantial +14.7 absolute point improvement over MiniCheck-7B (50.0 BAcc). The ThinknCheck-nothink ablation performs poorly (21.7 BAcc), confirming that reasoning drives this enhanced generalization. These results demonstrate that ThinknCheck handles claims requiring deeper understanding more effectively, with important implications for deployment in domain-shifting scenarios.

\paragraph{Preference optimization.}
Preference optimization with a simple two-term reward (format + class-weighted accuracy) underperforms supervised reasoning: GRPO from base yields 52.6 BAcc (below 55.7), and GRPO from an SFT warm start reaches 74.2 (below 78.1). A manual audit indicates GRPO gravitates toward a lexical-overlap shortcut (predicting \textsc{yes} when the claim reuses document phrasing). At 1B scale, explicit reasoning supervision is a more reliable path than zero-shot CoT or coarse preference rewards for grounded verification. Full GRPO details are in Appendix~\ref{app:grpo}.

\subsection{Cross-Benchmark Performance: GSMClaims and ThinknCheck-Science}
\begin{table}[h]
\centering
\small
\resizebox{\columnwidth}{!}{%
\begin{tabular}{lccc}
\toprule
\textbf{Model} & \textbf{LLMAggreFact} & \textbf{SciFact} & \textbf{GSMClaims} \\
               & \textbf{BAcc}          & \textbf{BAcc}    & \textbf{Acc}       \\
\midrule
MiniCheck-7B & 77.4 & 50.0 & 51.3 \\
ThinknCheck-nothink-1B & 57.5 & 21.7 & 49.4 \\
ThinknCheck-1B & 78.1 & 64.7 & 52.2 \\
ThinknCheck-Science-1B & \textbf{79.2} & \textbf{66.4} & \textbf{61.0} \\
\bottomrule
\end{tabular}
}
\caption{Performance across all benchmarks. ThinknCheck-Science, trained with additional scientific and arithmetic data, achieves the best performance on all three evaluations, including 61.0\% accuracy on GSMClaims (17\% relative improvement over ThinknCheck-1B).}
\label{tab:cross-benchmark}
\vspace{-1.0em}
\end{table}

We used the GSMClaims dataset (Section~\ref{sec:gsmclaims-creation}) to evaluate arithmetic reasoning. As expected, both ThinknCheck-1B (52.2\%) and MiniCheck-7B (51.3\%) find this task challenging, confirming that current datasets relying heavily on textual entailment are insufficient for numerical reasoning. ThinknCheck-Science (Table~\ref{tab:cross-benchmark}), trained with additional scientific and arithmetic data, achieves 61.0\% on GSMClaims---a 17\% relative improvement. It also improves on LLMAggreFact (79.2) and SciFact (66.4), indicating that specialized training enhances rather than compromises general verification.

\subsection{How much reasoning is enough? Reasoning Length vs. Accuracy}
\label{sec:length-v-bacc}
\begin{figure}[t]
  \centering
  \includegraphics[width=\linewidth]{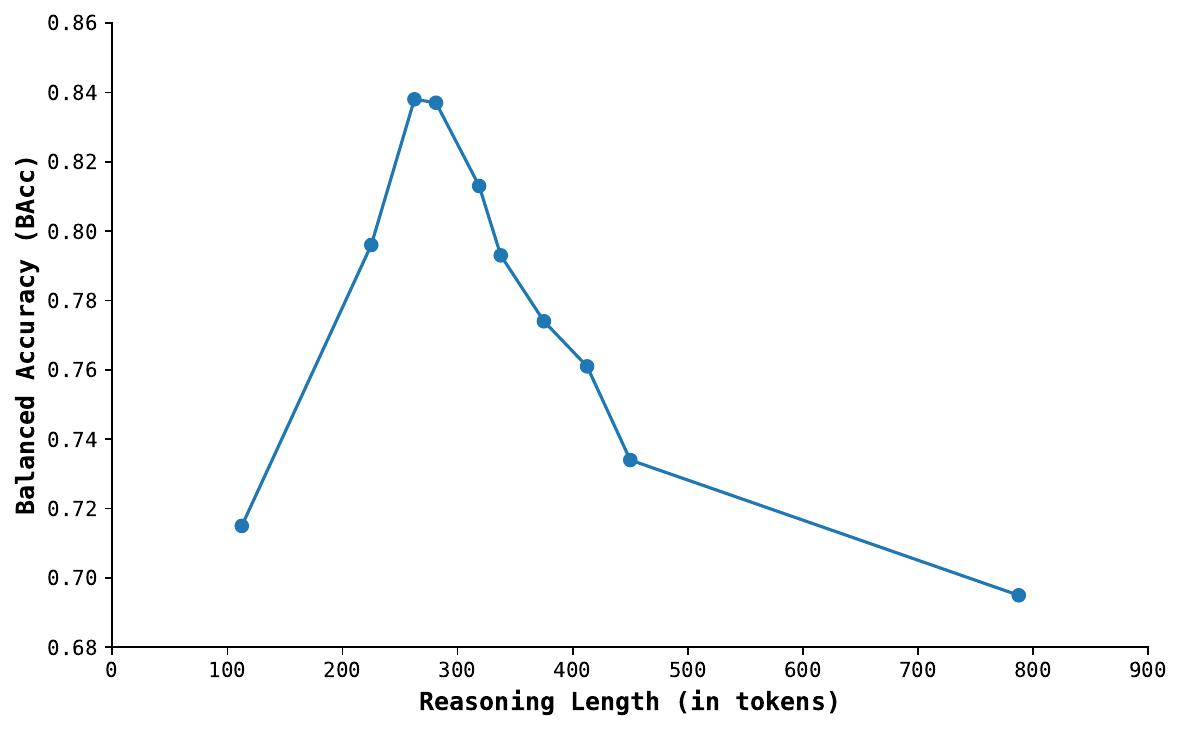}
  \caption{\textbf{Reasoning length vs.\ balanced accuracy (LLMAggreFact).}
  ThinknCheck-1B outputs are grouped into deciles by the token length of the \texttt{<REASONING>} span (Gemma tokenizer).
  BAcc peaks for mid-length rationales and drops for very short and very long chains. Short chains show recall $>$ precision;
  very long chains show the opposite trend.}
  \label{fig:length-bacc}
  \vspace{-0.2em}
\end{figure}

We study ThinknCheck-1B on the LLMAggreFact test set to relate the amount of generated reasoning to correctness: we tokenize the \texttt{<REASONING>} span with the Gemma3 tokenizer, bucket examples into ten equal-sized deciles by token length, and compute BAcc per decile (Figure~\ref{fig:length-bacc}). Accuracy follows an inverted-U: mid-length rationales yield the best BAcc, whereas very short and very long chains underperform. Short chains exhibit recall exceeding precision, consistent with liberal ``YES'' predictions triggered by shallow matches; at the longest lengths, recall drops below precision, reflecting conservative behavior and more false negatives. Length is not causal---longer chains co-vary with instance difficulty and the need to aggregate more evidence---but the pattern aligns with our error taxonomy (Section~\ref{sec:error-analysis}): terse chains co-occur with lexical-overlap false positives, and very long chains with over-cautiousness and insufficient aggregation. Together with the ablations in Section~\ref{sec:core-ablations}, the takeaway is that focused, succinct-but-substantive reasoning, rather than maximal length, is most reliable at 1B scale.

\section{Error Analysis}
\label{sec:error-analysis}

\begin{figure*}[t]
  \centering
  \begin{subfigure}[b]{0.32\textwidth}
      \includegraphics[width=\textwidth]{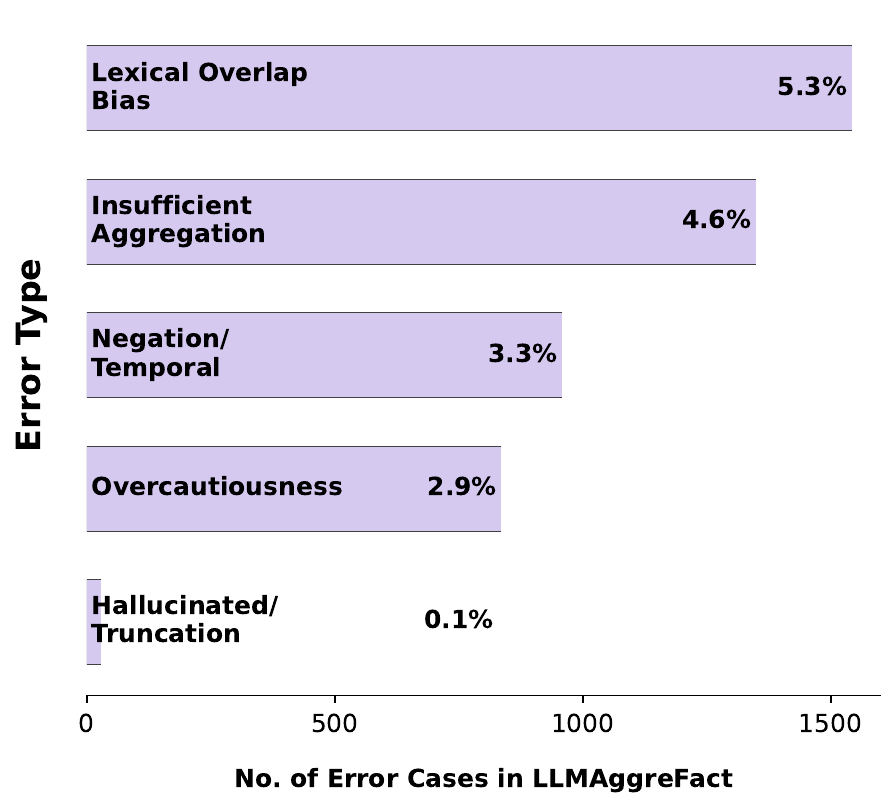}
      \caption{LLMAggreFact}
      \label{fig:error_llmaggrefact}
  \end{subfigure}
  \hfill
  \begin{subfigure}[b]{0.32\textwidth}
      \includegraphics[width=\textwidth]{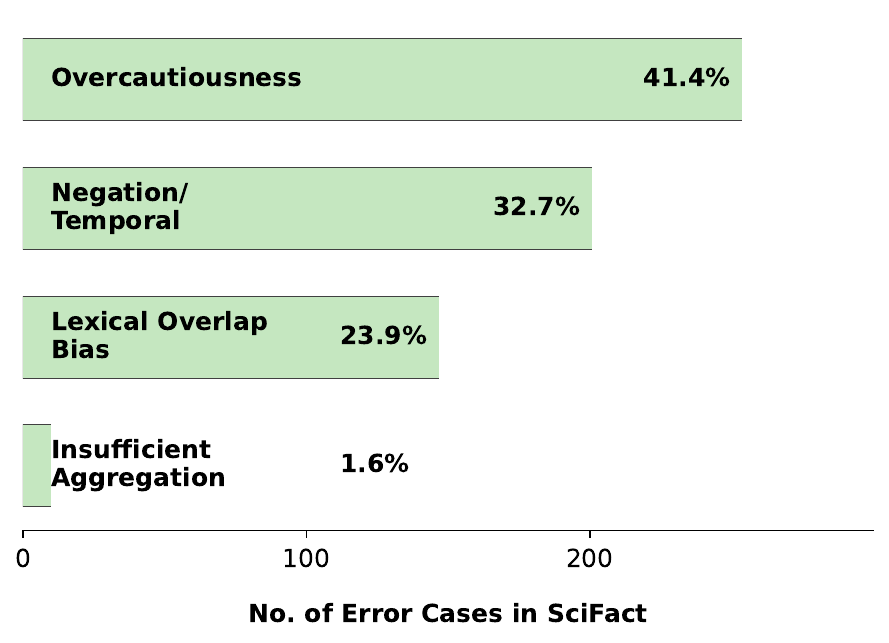}
      \caption{SciFact}
      \label{fig:error_scifact}
  \end{subfigure}
  \hfill
  \begin{subfigure}[b]{0.32\textwidth}
      \includegraphics[width=\textwidth]{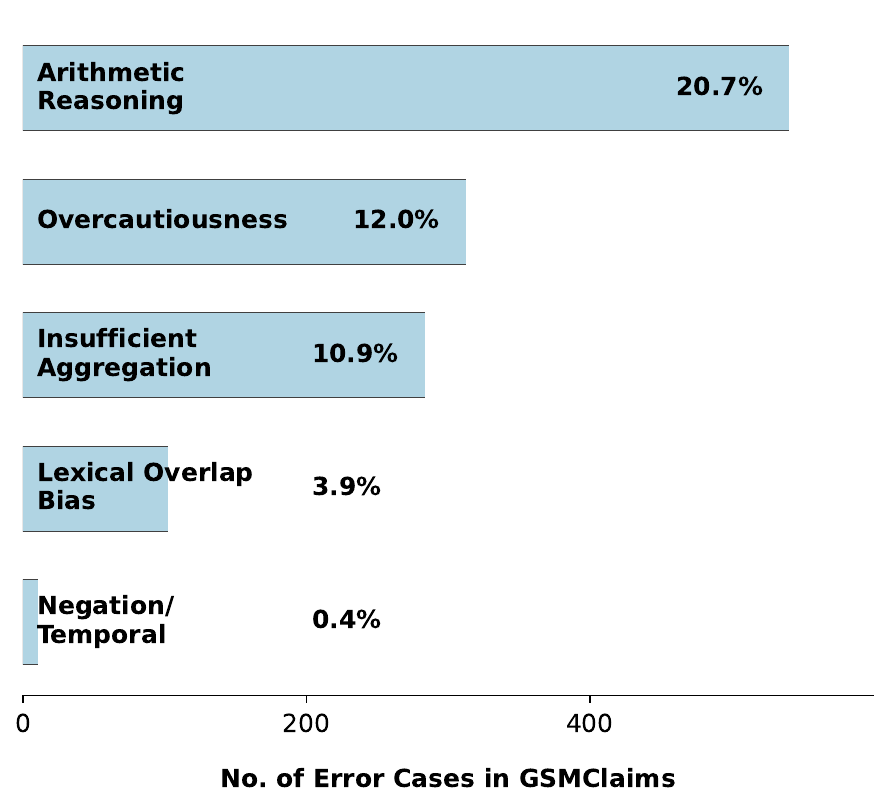}
      \caption{GSMClaims}
      \label{fig:error_gsm}
  \end{subfigure}
  \vspace{-0.7em}
  \caption{Distribution of error types on (a) LLMAggreFact, (b) SciFact, and (c) GSMClaims. Error profiles vary dramatically by domain: general claims are dominated by lexical overlap and aggregation failures; scientific claims by overcautiousness; mathematical claims by arithmetic reasoning errors.}
  \label{fig:combined_error_analysis}
  \vspace{-1em}
\end{figure*}

We analyzed ThinknCheck-1B errors on LLMAggreFact, SciFact, and GSMClaims using a unified taxonomy (Figure~\ref{fig:combined_error_analysis}). \textbf{Lexical Overlap Bias} was most prevalent in LLMAggreFact (5.3\%) but lower in GSMClaims (3.9\%). In GSMClaims, \textbf{Arithmetic Reasoning} errors dominated (43.2\% of errors). \textbf{Overcautiousness}, the leading error in SciFact (41.4\% of errors), reflects difficulty confirming complex scientific assertions. \textbf{Negation/Temporal} errors were significant in SciFact (32.8\%) but rare in GSMClaims (0.9\%). \textbf{Insufficient Aggregation} was critical in LLMAggreFact (4.6\%) where multi-hop synthesis is key.

GRPO optimization increased the share of lexical-overlap false positives: the policy frequently predicted \textsc{Yes} when the claim reused document phrases without verifying entailment. Consistent with Section~\ref{sec:length-v-bacc}, terse rationales coincide with lexical-overlap false positives, while very long rationales correlate with over-cautiousness and insufficient aggregation. These domain-specific error profiles suggest that no single mitigation strategy suffices; targeted approaches such as adversarial data mining and domain-specific prompting are needed.

\section{Conclusion}
\vspace{-0.5em}
We introduced ThinknCheck, a compact verifier trained to reason before deciding. At 1B parameters, supervised rationales raise performance on LLMAggreFact to 78.1 BAcc and improve out-of-domain generalization on SciFact (64.7 BAcc vs.\ 50.0 for MiniCheck-7B). Removing the reasoning step lowers BAcc by 20.6 points, and both zero-shot chain-of-thought on the base model and GRPO with a simple reward degrade accuracy under our setup. Analysis of model traces suggests that widely used verification datasets emphasize surface matching while under-testing evidence synthesis and arithmetic. Our GSMClaims benchmark and the domain-adapted ThinknCheck-Science address these gaps and yield gains across tasks, including 61.0\% accuracy on GSMClaims. A reasoning length study further indicates that mid-length rationales are most reliable for this setting, though length likely correlates with instance difficulty rather than causing accuracy changes. In sum, the results support small, reasoning-supervised verifiers as practical building blocks for grounded verification when compute, privacy, or latency constrain model size. Immediate next steps include constructing harder, balanced datasets that require multi-sentence synthesis and numerical reasoning, adding lightweight tool use for arithmetic, and learning adaptive reasoning budgets that keep rationales concise without sacrificing correctness.

\section{Limitations}
\label{sec:limitations}
ThinknCheck advances claim verification, and we identify several promising directions for future development. Our CoT result reflects a single prompt and decoding setup; while consistent with recent reports that CoT can hurt small models without adaptation, we did not exhaustively analyze all CoT prompt variations. Our GRPO setup uses a concise, two-term reward and a short training horizon (400 steps); alternative rewards (e.g., evidence-aware or contrastive), stronger KL or behavior-cloning constraints to the SFT policy, or longer training may change outcomes. We therefore interpret the GRPO results as a negative finding under a transparent, reproducible configuration rather than a general indictment of preference optimization. While the Gemma3-1B backbone supports a 32k-token context window, our experiments do not stress this limit: we evaluate single-document verification with relatively short inputs -- a limitation of current claim verification benchmarks. Extending ThinknCheck to truly long-context settings (multi-document packs~\cite{poli2023hyena,waleffe2024empirical}, $\geq$10k tokens) is left for future work. We also used a fixed maximum decoding budget for the \texttt{<REASONING>} span; making this budget adaptive could reduce both under- and over-thinking. Furthermore, performance on tasks like GSMClaims suggests that integrating external tools (e.g., calculators)~\cite{patil2024gorilla} is a key step for complex arithmetic reasoning. Finally, aligning with challenges in prior work~\cite{tang2024minicheck,zha2023alignscore}, calibrating output logits to serve as reliable confidence scores~\cite{liu2025uncertainty} remains an important area for ongoing investigation and future refinement of ThinknCheck.

All created datasets and models will be released under an Apache 2.0 license\footnote{URL withheld for blind review.}.

\section*{Acknowledgments}
This research was developed with funding from the Defense Advanced Research Projects Agency's (DARPA) SciFy program (Agreement No. HR00112520300). The views expressed are those of the author and do not reflect the official policy or position of the Department of Defense or the U.S. Government.

\bibliographystyle{splncs04}
\bibliography{custom}

\newpage
\appendix

\section{Prompt for generating LLMAggreFact-Think}
\label{appendix:think-prompt}
\begin{mdframed}[backgroundcolor=gray!10]
\noindent You are expert fact checker with a strong attention to detail
and access to a wealth of information. Given a document and a
claim, determine if the claim is entailed by the document,
only using the facts in the document.

\noindent Respond in the following format:
\begin{lstlisting}
<reasoning>
... // clear, but short description of your step by step
   // thinking to arrive at the entailment
   // keep the reasoning sentences separated by a newline.
</reasoning>
<entailment>
... // This is always a single word, either "YES" or "NO"
</entailment>
\end{lstlisting}
\end{mdframed}

\section{Hyperparameter Details}
\label{appendix:finetune-hyperparms}
For fine-tuning, we used LoRA \cite{hu2021lora} with rank=64, lora\_alpha=64, and a learning rate of 2e-4 scheduled linearly. We updated the query, key, value, and output projection layers, as well as MLP gate, up, and down projections. The fine-tuning was performed on an A100 GPU for 1 epoch, with 5 warmup steps, a batch size of 4 with 4 accumulated steps, and an 8bit-AdamW optimizer with a weight decay of 0.01.

\section{Finetuning Prompt for ThinknCheck}
\label{appendix:finetune-prompt-thinkncheck}
\begin{mdframed}[backgroundcolor=gray!10]
\noindent You are given a document and a claim. The document is
enclosed between \texttt{<DOCUMENT>} and \texttt{</DOCUMENT>}. The claim
is between \texttt{<CLAIM>} and \texttt{</CLAIM>}. Determine if the claim
is entailed by the document. Think about the problem
and provide your reasoning. Place the reasoning
between \texttt{<REASONING>} and \texttt{</REASONING>}. Then, provide
your entailment solution between \texttt{<SOLUTION>} and
\texttt{</SOLUTION>}. The entailment should be either a YES or a
NO.
\begin{lstlisting}
<DOCUMENT>
{document}
</DOCUMENT>
<CLAIM>
{claim}
</CLAIM>
<REASONING>
{reasoning}
</REASONING>
<SOLUTION>
{solution}
</SOLUTION>
\end{lstlisting}
\end{mdframed}

\section{Finetuning Prompt for ThinknCheck-nothink}
\label{appendix:finetuning-prompt-nothink}
\begin{mdframed}[backgroundcolor=gray!10]
\noindent You are given a document and a claim. The document is
enclosed between \texttt{<DOCUMENT>} and \texttt{</DOCUMENT>}. The claim
is between \texttt{<CLAIM>} and \texttt{</CLAIM>}. Determine if the claim
is entailed by the document. Provide your entailment
solution between \texttt{<SOLUTION>} and \texttt{</SOLUTION>}. The
entailment should be either a YES or a NO.
\begin{lstlisting}
<DOCUMENT>
{document}
</DOCUMENT>
<CLAIM>
{claim}
</CLAIM>
<SOLUTION>
{solution}
</SOLUTION>
\end{lstlisting}
\end{mdframed}

\section{GRPO Details}
\label{app:grpo}
We use TRL's \texttt{GRPOTrainer} with the Dr.~GRPO loss and GSPO. The reward is $R = 0.5 \cdot R_{\text{fmt}} + 1.0 \cdot R_{\text{acc}}$, where $R_{\text{fmt}} \in \{+1,-1\}$ checks for both \texttt{<REASONING>} and \texttt{<SOLUTION>} tags, and $R_{\text{acc}}$ applies class-weighted accuracy ($w_{\textsc{YES}}{=}0.6356$, $w_{\textsc{NO}}{=}2.3435$, scale ${\pm}4.0$). We draw 5 completions per prompt, cap prompts at 4{,}306 tokens and completions at 378 tokens, and train for one epoch (400 steps) with \texttt{adamw\_torch\_fused} (LR $5{\times}10^{-6}$, warmup 0.1, batch 4, gradient accumulation 4). LoRA adapters are initialized from an SFT-400 checkpoint. GRPO variants use the same I/O format and decoding defaults as SFT.

\section{Prompt to generate GSMClaims}
\label{appendix:GSMClaims-prompt}
\begin{mdframed}[backgroundcolor=gray!10]
\noindent Given an arithmetic problem and a solution, rewrite them as a document and a a pair of positive and negative claims such the positive claim is entailed by the document (after solving some arithmetic) and the negative claim is not entailed by the document (after solving some arithmetic). Produce your answer only as a JSON. Do not add anything before and after the JSON.
\end{mdframed}

\end{CJK*}

\end{document}